\documentclass{article} 
\usepackage{iclr2027_conference,times}

\usepackage{amsmath,amsfonts,bm}

\def\eqref#1{equation~\ref{#1}}

\def\1{\bm{1}}

\DeclareMathAlphabet{\mathsfit}{\encodingdefault}{\sfdefault}{m}{sl}
\SetMathAlphabet{\mathsfit}{bold}{\encodingdefault}{\sfdefault}{bx}{n}

\usepackage{hyperref}
\usepackage{url}
\usepackage{multirow}
\usepackage{booktabs}
\usepackage{graphicx}
\usepackage{wrapfig}
\usepackage[normalem]{ulem}

\usepackage{algorithm}
\usepackage{algpseudocode}
\usepackage{booktabs}
\usepackage{float}
\usepackage{wrapfig}

\floatstyle{ruled}
\restylefloat{algorithm}

\title{CapField-OPD: Learning Continuous Capability Fields via Joint-Anchored Multi-Teacher On-Policy Distillation for Flow Models}

\author{%
Pengyang Ling\textsuperscript{1,*},
Jiazi Bu\textsuperscript{2,*},
Yujie Zhou\textsuperscript{2,*},
Yibin Wang\textsuperscript{3}, 
Zeqiang Lai\textsuperscript{4},\\
{\bfseries Xiaoxiao Ma\textsuperscript{1},
Yi Jin\textsuperscript{1}, 
Huaian Chen\textsuperscript{1,\dag}, 
Yuhang Zang\textsuperscript{5,\dag}}\\
\textsuperscript{1}University of Science and Technology of China
\textsuperscript{2}Shanghai Jiao Tong University 
\textsuperscript{3}Fudan University \quad \\
\textsuperscript{4}The Chinese University of Hong Kong \quad
\textsuperscript{5}Shanghai Artificial Intelligence Laboratory\\
{\small *~Equal contribution. \quad \dag~Corresponding authors.}
}

\iclrfinalcopy 
\begin{document}

\maketitle
\fancyhead[L]{}                    

\vspace{-1.5em}
\begin{abstract}
Reward-specialized post-training produces strong experts for flow-based generative models, while multi-teacher on-policy distillation (OPD) consolidates their capabilities into a single student. Existing methods, however, route each prompt to a single teacher according to its semantic category, implicitly binding the desired capability to prompt content. This coupling makes capability invocation vulnerable to prompt perturbations and prevents users from explicitly adjusting the strength of the desired capability at inference time. In this work, we introduce CapField-OPD, an OPD framework that integrates multiple teachers into a continuous capability field through explicit capability coordinates. We use teacher models as anchors to construct this field, with the coordinates determining how their outputs are combined. Each capability configuration thus receives a unique supervision target, and capability control no longer depends on prompt semantics. Since the training anchors may not be optimal at inference time, we further profile the learned field on a small calibration set. The coordinate with the highest mean reward serves as the recommended default, while coordinates that are frequently optimal offer a promising candidate set for test-time scaling. Extensive experiments on compositional generation, text rendering, and visual aesthetics demonstrate that CapField-OPD consolidates multiple specialized teachers into a single student while preserving or surpassing their performance, reliably invokes the desired capabilities under semantics-preserving prompt variations, and supports continuous capability control and coordinate-based test-time scaling.
\end{abstract}

\section{Introduction}

Flow-matching models~\citep{esser2024scaling, lipman2022flow,liu2022flow}
have recently become a popular framework for image generation~\citep{flux2024, team2025zimage, wu2025qwen}, 
in which a learned velocity field can transport Gaussian noise to data samples via iterative denoising. 
Reward-based post-training~\citep{liu2025flow, xue2025dancegrpo, Zhou_2026_CVPR} 
further improves specific generation capabilities, 
such as text-rendering accuracy,
compositional fidelity, and visual aesthetics,
producing strong and diverse domain experts.
However, training and deploying a separate model for each capability is costly.
Multi-teacher on-policy distillation (OPD)~\citep{fang2026flow} addresses this problem by consolidating
reward-specialized teachers into a single student, 
which is supervised by teachers' outputs evaluated along student-generated trajectories.

\begin{figure*}[t]
    \centering
    \includegraphics[width=\textwidth]{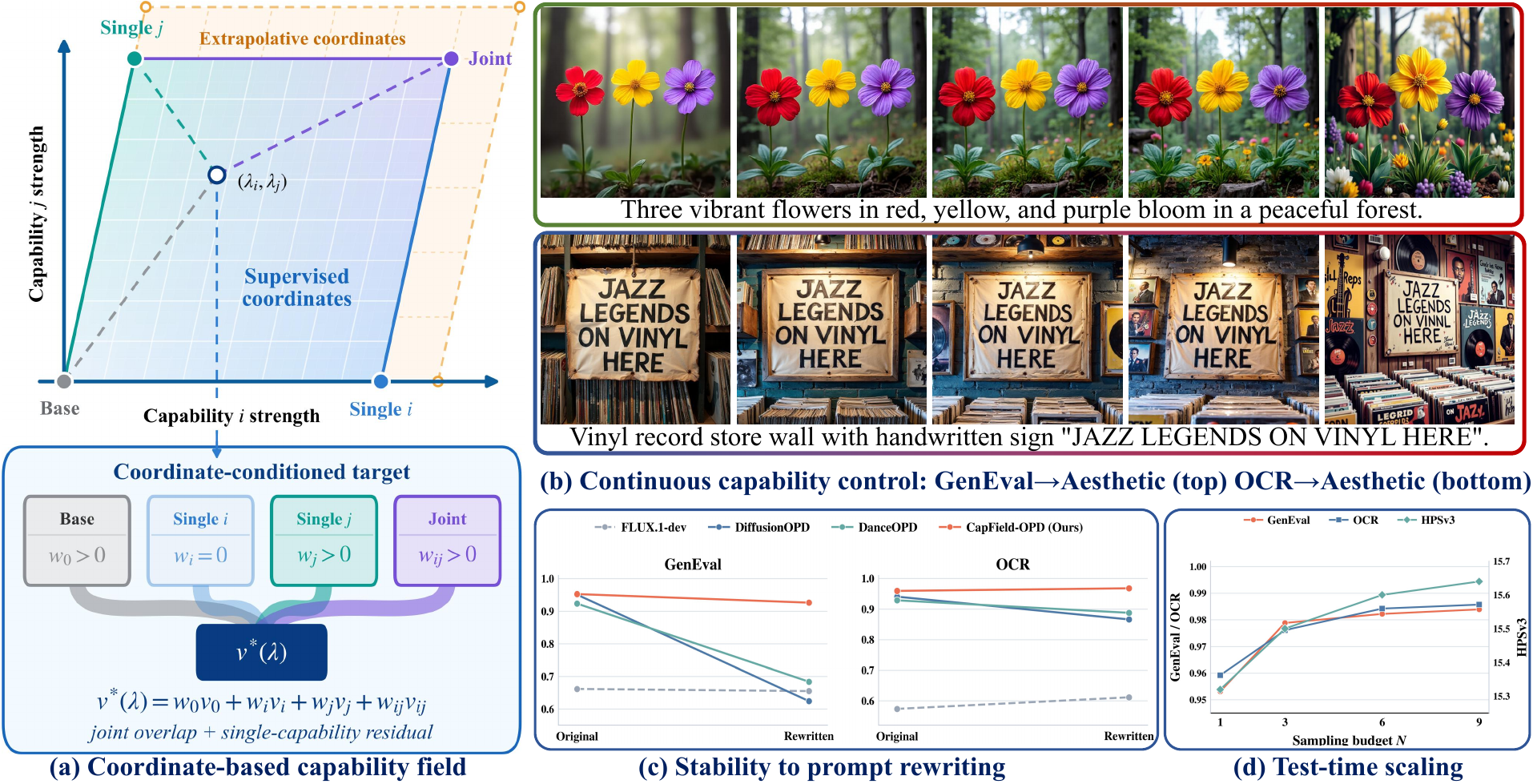}
    \vspace{-2em}
    \caption{
Overview of CapField-OPD.
(a) Teacher anchors define a continuous capability field controlled by
explicit, potentially extrapolative coordinates.
(b-d) The learned field supports continuous control, robust capability
invocation under prompt rewriting, and coordinate-based test-time scaling.
}
    \label{fig:overview}
    \vspace{-1.8em}
\end{figure*}

Nevertheless, existing flow-based multi-teacher OPD
methods~\citep{fang2026flow, li2026diffusionopd, zhou2026danceopd} typically
split training prompts by task type and assign each subset to one teacher via
semantics-driven hard routing. Without an explicit capability signal, the
student must infer the desired capability category from prompt semantics,
binding capability intent to prompt content. Capability activation thus
becomes sensitive to prompt wording and style, especially on
out-of-distribution prompts, and users cannot explicitly control capabilities
at inference time. As shown in Fig.~\ref{fig:overview}(b), for the same prompt
describing a handwritten sign reading ``JAZZ LEGENDS ON VINYL HERE,'' one user
may prioritize accurate text rendering, another may prefer visual aesthetics,
while a third may require both with different strengths. A semantic router
cannot distinguish these intentions from the prompt alone. When a prompt
requires multiple capabilities, naively activating multiple teachers yields
conflicting velocity targets under the same student condition. These
limitations point to a key issue: capability intent requires explicit
representation and control.

To this end, we propose CapField-OPD, built on the principle that desired
capabilities should be explicitly specified rather than implicitly inferred.
Specifically, we condition the student
on an explicit capability coordinate whose axes control capability strengths,
and reinterpret the base, single-capability, and joint-capability teachers as
anchors of a shared capability field. A joint-anchored target field combines
their outputs through coordinate-dependent activation weights. For coordinates
activating multiple capabilities, the corresponding joint-capability teacher
supplies the shared capability components, and remaining activation is
allocated to the single-capability teachers, so the field exactly preserves
each anchor's capability while providing a continuous velocity target over the
capability space. The student can thus invoke desired capabilities even on
boundary or out-of-distribution prompts where semantic routing may fail
(Fig.~\ref{fig:overview}(c)). Coordinate conditioning also adds an
inference-time scaling axis. Although supervision spans a bounded range,
varying the coordinates teaches the student how its velocity field changes as
capabilities strengthen or combine. The teacher anchors therefore define
reference states rather than hard performance limits. As shown in
Fig.~\ref{fig:overview}(a), extending coordinates beyond the training range
continues the learned capability response, and this bounded extrapolation can
yield higher rewards than the teacher anchors. To identify reliable
extrapolative settings, we profile in-range and extrapolative coordinates on a
small calibration set of prompt--noise pairs. The coordinate with the highest
mean reward serves as the default, and frequently optimal coordinates provide
candidates for prompt-specific search, enabling coordinate-based test-time
scaling (Fig.~\ref{fig:overview}(d)).

Our contributions are as follows:
(1) We identify semantic-capability entanglement of hard routing as a central
limitation of multi-teacher OPD.
(2) We propose CapField-OPD, which learns a continuous capability field
through explicit capability conditioning and joint-anchored field supervision,
with capability landscape profiling for default-coordinate selection and
test-time search.
(3) Experiments show CapField-OPD consolidates reward-specialized experts,
enhances robustness to prompt perturbations, and supports continuous
capability control and coordinate-based test-time scaling.

\section{Related work}

\subsection{Preference alignment in flow models}
Aligning pretrained flow models with human preferences~\citep{liu2025improving, xue2025dancegrpo,bu2026sparse} 
has emerged as an effective post-training strategy for 
adapting them to downstream objectives.
Existing approaches can be broadly divided into offline preference learning
and online reward optimization. 
Offline DPO-style methods~\citep{wallace2024diffusion,yang2024using} 
translate preferred–-dispreferred comparisons into flow-matching objectives~\citep{liu2025improving},
thereby avoiding explicit reward-model training and online rollouts.
Online methods~\citep{liu2025flow, ling2026pave, li2026tmpo}, in contrast, 
formulate the sampling trajectory as a Markov decision process (MDP) 
and interleave trajectory sampling and reward scoring with policy updates. 
Specifically, the PPO-style methods (DPOK~\citep{fan2023dpok}, DDPO~\citep{black2023training}) apply policy gradients along denoising trajectories, 
whereas GRPO-style approaches~\citep{liu2025flow, li2025mixgrpo,wang2025coefficients} convert an ordinary differential equation (ODE) sampling
into an equivalent stochastic differential equation (SDE) 
and optimize the sampling distribution using estimated relative advantages. 
Both of them rely on tractable stochastic transition densities 
and perform on-policy optimization along the sampled trajectories. 
In comparison, DiffusionNFT~\citep{zheng2026diffusionnft}
and Advantage Weighted Matching (AWM)~\citep{xue2025advantage} develop a different approach.
They collect and score generated images, 
and then re-noise them for reward-derived weighted optimization, 
achieving substantially faster convergence.
Collectively, these techniques improve the performance of preference alignment toward specific reward functions.

\subsection{On-policy distillation in flow models}
Unlike conventional distillation methods~\citep{yin2024improved, yin2024one, cheng2025twinflow,chen2025sana} 
that primarily aim to reduce sampling steps, 
on-policy distillation (OPD)~\citep{fang2026flow, li2026diffusionopd, zhou2026danceopd} has recently emerged as an effective approach 
for consolidating the capabilities of multiple teachers into a single student model. 
Instead of relying on fixed data or teacher-generated results, 
OPD samples trajectories from the current student and queries teachers at student-visited states,
reducing exposure bias while providing dense velocity-field supervision. 
Specifically, Flow-OPD~\citep{fang2026flow} and DiffusionOPD~\citep{li2026diffusionopd} independently train reward-specialized teachers 
and consolidate their capabilities into a unified student through task routing and trajectory-level matching. 
Subsequent works extend OPD in several directions.
For example, DreOPD~\citep{lin2026dreopd} moves beyond direct teacher imitation through degraded-reference velocity extrapolation, 
while Any-OPD~\citep{fu2026any} enables distillation between heterogeneous teacher--student pairs through representation-space alignment.
In addition, CFG-OPD~\citep{li2026rethinking} separately constrains the positive prediction and the CFG direction, reducing sensitivity to guidance scales. 
Although these methods improve OPD for flow models, 
existing multi-teacher OPD methods still rely on semantics-driven hard routing, 
assigning each trajectory to one expert.
This entangles prompt content with capability intent,
limiting capability composition and strength control 
and causing conflicting supervision for multi-capability or boundary prompts.

\section{Method}

\subsection{Preliminaries}

\paragraph{Flow Matching Models.} Flow matching~\citep{lipman2022flow,
liu2022flow} learns a time-dependent velocity field that transports Gaussian
noise to clean data. Given a paired sample $(\mathbf{x},c)\sim
p_{\mathrm{data}}$, Gaussian noise
$\boldsymbol{\epsilon}\sim\mathcal{N}(\mathbf{0},\mathbf{I})$, and a timestep
$t\sim\mathcal{U}[0,1]$, the interpolated noisy state is
$\mathbf{x}_t=(1-t)\mathbf{x}+t\boldsymbol{\epsilon}$, where
$\mathbf{x}_0=\mathbf{x}$ is the clean sample and
$\mathbf{x}_1=\boldsymbol{\epsilon}$ is pure noise. Differentiating the path
with respect to $t$ gives the target velocity
\begin{equation}
    v_t
    =
    \frac{\mathrm{d}\mathbf{x}_t}{\mathrm{d}t}
    =
    \boldsymbol{\epsilon}-\mathbf{x}.
\end{equation}
The flow model $v_\theta(\mathbf{x}_t,t,c)$ is then trained with the flow
matching loss
\begin{equation}
    \mathcal{L}_{\mathrm{FM}}(\theta)
    =
    \mathbb{E}_{(\mathbf{x},c),\boldsymbol{\epsilon},t}
    \left[
    \left\|
    v_\theta(\mathbf{x}_t,t,c)-v_t
    \right\|_2^2
    \right],
\end{equation}
where $c$ is an optional condition such as text or image.

\paragraph{On-Policy Distillation.}
On-policy distillation (OPD)~\citep{li2026diffusionopd} aims to transfer the capabilities of frozen teachers to a student by imitating the teacher's behavior on student-visited states, thereby offering supervision that adapts to the student's evolving distribution. Let $v_\phi$ and $v_\theta$ denote the teacher and student velocity fields, respectively, and let
$\tau=\{\mathbf{x}_{t_i}\}_{i=0}^{M}$
be the trajectory sampled by the student model; the OPD objective is defined as: 
\begin{equation}
    \mathcal{L}_{\mathrm{OPD}}(\theta)
    =
    \mathbb{E}_{c,\tau\sim p_\theta(\tau\mid c)}
    \left[
    \sum_{i=0}^{M-1}
    w(t_i)
    \left\|
    v_\theta(\mathbf{x}_{t_i},t_i,c)
    -
    v_\phi(\mathbf{x}_{t_i},t_i,c)
    \right\|_2^2
    \right],
\end{equation}
where $w(t_i)$ is a timestep-dependent weight. When the student and teacher models share the same per-step covariance, this velocity-matching loss is equivalent to minimizing the per-step KL divergence between the two models (omit the time-dependent coefficient):
\begin{equation}
    D_{\mathrm{KL}}
    \!\left(
    \pi_\theta(\cdot\mid\mathbf{x}_{t_i})
    \,\|\,  
    \pi_\phi(\cdot\mid\mathbf{x}_{t_i})
    \right)
    \propto
    \left\|v_\theta-v_\phi\right\|_2^2.
\end{equation}

\subsection{Observation}
\label{sec:overview}

Suppose that $K$ reward-specialized teachers encode capabilities such as
compositional fidelity, text rendering, and visual aesthetics. To mitigate
cross-teacher supervision conflicts, existing multi-teacher OPD methods
partition the training prompts into non-overlapping subsets
$\{\mathcal{D}_k\}_{k=1}^{K}$ and assign each subset to one teacher:
\begin{equation}
v^{\star} = v_{\phi_k}, \qquad k = \rho(c) \ \ \text{for}\ \ c \in \mathcal{D}_k ,
\end{equation}
where $\rho$ is the fixed task-based routing rule and $v_{\phi_k}$ denotes
the $k$-th teacher. The selected index $k$ determines the supervision
target but is not provided to the student, forcing it to infer the
intended capability solely from $c$ and thereby entangling prompt
semantics with capability activation.

\begin{wrapfigure}{r}{0.5\textwidth}
    \centering
    \vspace{-1em}
    \includegraphics[width=\linewidth]{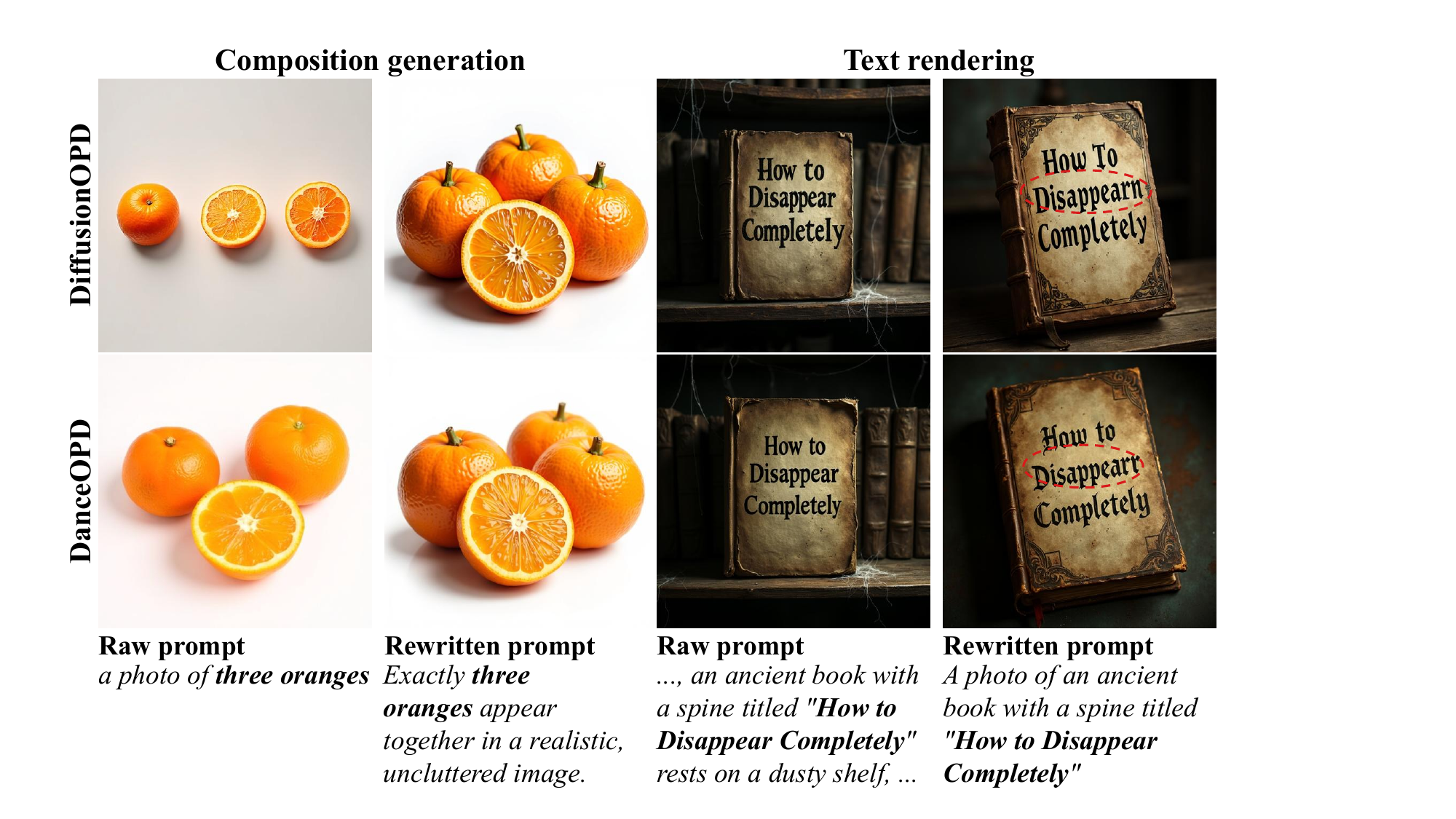}
    \vspace{-2em}
    \caption{
        Visualization under prompt rewriting.
    }   
    \label{fig:robustness}  
    \vspace{-1em}
\end{wrapfigure}

This design leads to two limitations. First, capability activation is
sensitive to linguistic variation: when the wording or style of a prompt
deviates from the routed training subsets, the prompt may activate a
different expert behavior even if its visual intent is nearly identical
(Fig.~\ref{fig:overview}(c) and Fig.~\ref{fig:robustness}). This resembles
the linguistic hacking observed in reward-based post-training~\cite{wang2026promptrl}, but is more
pronounced here because a perturbation can switch which capability is
activated rather than only changing its strength. Second, semantic routing
fixes both capability selection and activation strength in the prompt, so
users cannot adjust the strength of a single capability or invoke multiple
capabilities together for the same prompt.
These limitations arise from entangling semantic content with capability
intent and motivate an explicit capability-conditioning mechanism that
decouples \emph{what to generate} from \emph{which capabilities to
activate} and \emph{at what strength}, so that capability control no
longer depends on prompt wording and users can combine and adjust
capabilities at inference time.

\subsection{Joint-Anchored Explicit Capability Field}
\label{sec:capability_field}

We parameterize capability intent by $\boldsymbol{\lambda}\in[0,1]^K$, where
$\lambda_i$ continuously controls capability $i$, and multiple nonzero
entries indicate a request for capability combination.
The student $v_\theta(\mathbf{x}_t,t,c,\boldsymbol{\lambda})$ receives the
coordinate through a small projection added to its global conditioning, i.e.,
\begin{equation}
    \mathbf{h}(t,c,\boldsymbol{\lambda})
    =\mathbf{h}(t,c)+W_2\operatorname{SiLU}(W_1\boldsymbol{\lambda}).
    \label{eq:capability_conditioning}
\end{equation}
The projection is bias-free, and $W_2$ is initialized to zero. This separates
capability control from the text prompt without changing the text encoder
or transformer blocks.

Let $v_0$ denote the frozen base model without post-training, $v_i$ the
expert specialized for capability $i$, and $v_{ij}$ the joint teacher
obtained by jointly optimizing the objectives for capabilities $i$ and $j$
during post-training. Consider a capability coordinate
$\boldsymbol{\lambda}$ that is nonzero only on a supported pair $(i,j)$.
The corresponding target velocity is defined by the piecewise-linear
capability field
\begin{equation}
    v^*(\boldsymbol{\lambda})
    =(1-\lambda_i-\lambda_j+\lambda_{\min})v_0
    +(\lambda_i-\lambda_{\min})v_i
    +(\lambda_j-\lambda_{\min})v_j
    +\lambda_{\min} v_{ij},
    \label{eq:capability_field}
\end{equation}
where $\lambda_{\min}=\min(\lambda_i,\lambda_j)$ and, all velocities at
the same student-visited state $(\mathbf{x}_t,t,c)$, omitted for
clarity. The overlap $\lambda_{\min}$ of the two requests goes to the
joint teacher, and each single expert covers only the excess of its own
coordinate. Specifically, the corner $(\lambda_i,\lambda_j)=(0,0)$ recovers the
base model, $(1,0)$ and $(0,1)$ recover the two individual experts, and
$(1,1)$ recovers the joint teacher (Fig.~\ref{fig:overview}(a)). Anchored at
these four corners, the field is continuous and leaves each teacher
exactly reachable from any direction, so composing capabilities does not
degrade individual performance.
More broadly, Eq.~\ref{eq:capability_field} is not tied to a specific pair but provides a general rule for composing any two interacting capabilities,
which effectively alleviates conflicts between capabilities.
The joint teacher $v_{ij}$ contributes to the target only when both coordinates are
nonzero, since its coefficient $\lambda_{\min}$ vanishes otherwise. At an
axis-aligned coordinate such as $(\lambda_i,0)$, the target reduces to
interpolation between the base model and a single expert,
$v^*=(1-\lambda_i)v_0+\lambda_i v_i$, and single-capability supervision
never involves a joint teacher. The sampling region of each pair is therefore chosen according to its available anchors. For a pair with a joint teacher, coordinates are sampled over the full unit square, so that combined activations are also supervised. For a pair without a joint teacher, only axis-aligned coordinates are sampled, and the joint term is never used.

The training objective adopts the standard OPD loss, with both the
student and target velocity conditioned on the capability coordinates, i.e.,
\begin{equation}
    \mathcal{L}_{\mathrm{OPD}}(\theta)
    =\mathbb{E}\!\left[
    \left\|
    v_\theta(\mathbf{x}_t,t,c,\boldsymbol{\lambda})
    -v^*(\mathbf{x}_t,t,c,\boldsymbol{\lambda})
    \right\|_2^2
    \right].
    \label{eq:capfield_opd}
\end{equation}
As the target velocity varies with $\boldsymbol{\lambda}$, each requested
capability configuration is matched with its unique target, thus
the same prompt can be supervised by different teachers or their
combinations under different capability coordinates, without introducing
inconsistent targets for the same input.

\subsection{Capability Landscape Profiling}
\label{sec:capability_peak_profiling}

The value $\lambda_i=1$ marks the training anchor of capability $i$,
but does not necessarily yield the strongest student response.
We therefore profile the reward landscape of the learned capability
field to identify the coordinate with the highest mean reward and
measure how often each coordinate is optimal across prompt--noise pairs.
Given a capability set $S$, we fix all coordinates outside $S$ to zero
and construct a probe grid
$\Lambda_{\mathrm{probe}}^{S}
=\Lambda_{\mathrm{in}}^{S}\cup\Lambda_{\mathrm{ext}}^{S}$,
covering the training range and a bounded extrapolation region.
Every coordinate is evaluated on the same calibration bank
$\mathcal{B}_{\mathrm{cal}}^{S}=\{(c_m,z_m)\}_{m=1}^{M}$,
where $c_m$ is a training prompt and $z_m$ is its fixed initial noise.
The resulting reward records are
\begin{equation}
r_m(\boldsymbol{\lambda})
=
q_S\!\left(G_\theta(z_m,c_m,\boldsymbol{\lambda})\right),
\quad
m=1,\ldots,M,\quad
\boldsymbol{\lambda}\in\Lambda_{\mathrm{probe}}^{S},
\label{eq:profiling_records}
\end{equation}
where $q_S$ is the corresponding reward function for capability set $S$ (can be a weighted objective under multiple capabilities), and $G_\theta(\cdot)$ is the image generator.
Reusing the same prompt--noise pairs ensures that only the
capability coordinates vary across evaluations.
From these records, we compute the mean reward and the empirical
frequency of optimality at each coordinate:
\begin{equation}
\bar r_S(\boldsymbol{\lambda})
=
\frac{1}{M}\sum_{m=1}^{M}r_m(\boldsymbol{\lambda}),
\qquad
\widehat P_S(\boldsymbol{\lambda})
=
\frac{1}{M}\sum_{m=1}^{M}
\frac{\mathbf{1}\{\boldsymbol{\lambda}\in\Lambda_m^\star\}}
     {|\Lambda_m^\star|},
\label{eq:landscape_statistics}
\end{equation}
where $\bar r_S(\boldsymbol{\lambda})$ is the average reward of
coordinate $\boldsymbol{\lambda}$ over the $M$ pairs, $\widehat
P_S(\boldsymbol{\lambda})$ is the fraction of pairs for which
$\boldsymbol{\lambda}$ is optimal, with ties split evenly among the
optimal coordinates, and $\Lambda_m^\star
=\arg\max_{\boldsymbol{\lambda}\in\Lambda_{\mathrm{probe}}^{S}}
r_m(\boldsymbol{\lambda})$ is the set of optimal coordinates for the
prompt-noise pair $(c_m,z_m)$.
These two statistics measure different things: a coordinate that is best
on many pairs can still score poorly on the rest ( $\widehat P_S$
can be high while $\bar r_S$ stays moderate), and a coordinate can also
score well on every pair without ever being the best (the highest $\bar r_S$
can correspond to low $\widehat P_S$). We therefore define: 
\begin{equation}
\widehat{\boldsymbol{\lambda}}_{\mathrm{peak}}^{S}
=
\arg\max_{\boldsymbol{\lambda}\in\Lambda_{\mathrm{probe}}^{S}}
\bar r_S(\boldsymbol{\lambda}),
\qquad
\Lambda_{\mathrm{search}}^{S}
=
\left\{
\boldsymbol{\lambda}\in\Lambda_{\mathrm{probe}}^{S}:
\widehat P_S(\boldsymbol{\lambda})>0
\right\},
\label{eq:peak_and_search}
\end{equation}
where $\widehat{\boldsymbol{\lambda}}_{\mathrm{peak}}^{S}$ is the
recommended coordinate, and
$\Lambda_{\mathrm{search}}^{S}$ collects every
coordinate that is optimal for at least one prompt-noise pair, offering the candidate
set for test-time search. Since $\widehat P_S$ sums to one and is nonzero exactly on
$\Lambda_{\mathrm{search}}^{S}$, it can be used directly as the sampling distribution for test-time scaling.
The performance gain of coordinate extrapolation can thus be quantified as follows:
\begin{equation}
\Delta_{\mathrm{ext}}^{S}
=
\max_{\boldsymbol{\lambda}\in\Lambda_{\mathrm{ext}}^{S}}
\bar r_S(\boldsymbol{\lambda})
-
\max_{\boldsymbol{\lambda}\in\Lambda_{\mathrm{in}}^{S}}
\bar r_S(\boldsymbol{\lambda}).
\label{eq:extrapolation_difference}
\end{equation}

\subsection{Test-Time Scaling over Capability Coordinates}
\label{sec:test_time_scaling}

Although the default coordinate
$\widehat{\boldsymbol{\lambda}}_{\mathrm{peak}}^{S}$
achieves the highest mean reward during profiling, the optimal
coordinate can vary across prompt--noise pairs. We therefore
perform test-time search over capability coordinates to improve
generation quality for a given initial noise.
Specifically, we draw $N$ coordinates from
$\Lambda_{\mathrm{search}}^{S}$ without replacement, with
probability proportional to $\widehat P_S$.
Each coordinate produces a candidate record
$(\mathbf{x}_n,r_n,\boldsymbol{\lambda}_n)$:
\begin{equation}
\mathcal{C}_N
=
\left\{
(\mathbf{x}_n,r_n,\boldsymbol{\lambda}_n)
\right\}_{n=1}^{N},
\qquad
\mathbf{x}_n=G_\theta(z,c,\boldsymbol{\lambda}_n),
\quad
r_n=q_S(\mathbf{x}_n),
\label{eq:test_time_candidates}
\end{equation}
where the same initial noise $z$ is used for all candidates,
so that only the capability coordinates vary.
We then select the record with the highest reward:
\begin{equation}
(\mathbf{x}^{*},r^{*},\boldsymbol{\lambda}^{*})
\in
\arg\max_{(\mathbf{x},r,\boldsymbol{\lambda})\in\mathcal{C}_N}
r.
\label{eq:test_time_selection}
\end{equation}
Such a search explores different capability
configurations for the same initial noise, allowing
higher-reward outputs to be obtained without resampling
random seeds.

\section{Experiments}
\paragraph{Tasks and datasets.}
Following prior work~\citep{fang2026flow}, we investigate three capabilities:
compositional generation, text rendering, and visual aesthetics.
For compositional generation and text rendering, we use the training and
test splits released by Flow-GRPO~\citep{liu2025flow} for
GenEval~\citep{ghosh2023geneval} and OCR, respectively; for visual
aesthetics, prompts are drawn from the HPD~\citep{wu2023human} dataset.
Compositional generation and text rendering are evaluated with the GenEval
and OCR rewards, while the aesthetic objective combines
HPSv3~\citep{ma2025hpsv3}, CLIP~\citep{radford2021learning}, and
PickScore~\citep{kirstain2023pick} with equal weights
$1{:}1{:}1$.
For capability landscape profiling, the
calibration bank is drawn from the training split with 200 prompts per capability set, so that coordinate selection
never touches testing data.

\paragraph{Teacher construction.}
We train five teacher models using Flow-GRPO-Fast~\citep{liu2025flow}.
The three single-capability teachers are trained on the corresponding
prompts and rewards above.
We additionally train two joint-capability teachers, GenEval+aesthetics on
GenEval prompts and OCR+aesthetics on OCR prompts, where every generated
sample is evaluated by the task reward together with HPSv3, CLIP, and
PickScore at a coefficient ratio of $3{:}1{:}1{:}1$.
We do not build a GenEval+OCR teacher because the two prompt sets have
incompatible formats: GenEval prompts are template-based scene
descriptions, whereas OCR prompts must contain a quoted string to render,
so no natural prompt requires both capabilities.
Aesthetics imposes no constraint on prompt format and thus pairs with
either task; the joint teachers directly learn the combination of task
correctness and visual quality, providing anchors for coordinates where
both capabilities are activated.

\paragraph{Baselines.}
We use FLUX.1-dev~\citep{flux2024} as the backbone for all teacher and
student models and apply LoRA with $r=64$ and $\alpha=128$.
Rollout and distillation use 10 sampling steps; evaluation uses 28 steps,
and all images are generated at a resolution of $512\times512$.
We compare CapField-OPD with the three single-capability teachers, a
multi-task teacher trained through multi-objective reinforcement
learning~\citep{xue2025dancegrpo}, DiffusionOPD~\citep{li2026diffusionopd},
and DanceOPD~\citep{zhou2026danceopd}; all OPD methods use the same teacher
models for a fair comparison.
More implementation details can be found in the Appendix.


\begin{table}[t]
\centering
\caption{
Quantitative comparison. \emph{Single-only} and \emph{Single+Joint} distill from the three
single-capability teachers without and with the two joint-capability
teachers, respectively. The two CapField-OPD modes share one student
model and differ only in their inference coordinates. Bold and
underlined values denote the best and second-best results among unified models. Since aesthetic prompts contain no composition target or text-rendering target, the joint mode does not apply, and the corresponding entries are marked ``--''.
}
\label{tab:main_results}
\setlength{\tabcolsep}{3.2pt}
\renewcommand{\arraystretch}{1.12}
\resizebox{\textwidth}{!}{%
\begin{tabular}{lccccccccccc}
\toprule
\multirow{2}{*}{Method}
& \multicolumn{4}{c}{Composition-task prompts}
& \multicolumn{4}{c}{Text-rendering prompts}
& \multicolumn{3}{c}{Aesthetic prompts} \\
\cmidrule(lr){2-5}
\cmidrule(lr){6-9}
\cmidrule(lr){10-12}
& GenEval 
& HPSv3 
& CLIP 
& PickScore 
& OCR 
& HPSv3 
& CLIP 
& PickScore 
& HPSv3 
& CLIP 
& PickScore  \\
\midrule

FLUX.1-dev
& 0.6616 & 8.65 & 0.3966 & 23.40
& 0.5735 & 13.00 & 0.4466 & 22.91
& 13.28 & 0.3868 & 22.58 \\

\midrule
GenEval teacher
& 0.9417 & 9.16 & 0.4092 & 23.27
& 0.6170 & 13.43 & 0.4585 & 22.97
& 13.49 & 0.3946 & 22.47 \\

OCR teacher
& 0.7166 & 8.90 & 0.4030 & 23.59
& 0.9417 & 12.74 & 0.4567 & 22.88
& 13.57 & 0.3834 & 22.64 \\

Aesthetic teacher
& 0.3228 & 10.62 & 0.4117 & 24.04
& 0.4824 & 15.14 & 0.4688 & 23.95
& 15.23 & 0.4181 & 23.68 \\

GenEval+Aesthetic teacher
& 0.9253 & 11.75 & 0.4210 & 24.11
& 0.6235 & 14.74 & 0.4525 & 23.32
& 14.65 & 0.3968 & 22.71 \\

OCR+Aesthetic teacher
& 0.6041 & 9.46 & 0.4019 & 23.59
& 0.9031 & 14.40 & 0.4631 & 23.53
& 14.50 & 0.3945 & 23.06 \\

\midrule
Multi-task GRPO
& 0.8608 & 9.43 & \underline{0.4196} & \underline{23.89}
& 0.9369 & 13.20 & \textbf{0.4638} & 23.19
& 14.32 & 0.4024 & 23.15 \\

DiffusionOPD (Single-only)
& \underline{0.9514} & 8.37 & 0.4167 & 23.02
& \underline{0.9409} & 12.76 & 0.4574 & 22.90
& 15.25 & 0.4174 & \textbf{23.72} \\

DiffusionOPD (Single+Joint)
& 0.9434 & 10.96 & 0.4188 & 23.58
& 0.9229 & 13.86 & \underline{0.4632} & \underline{23.36}
& 15.25 & \underline{0.4180} & 23.70 \\

DanceOPD (Single-only)
& 0.9239 & 7.72 & 0.4092 & 22.97
& 0.9289 & 12.87 & 0.4579 & 22.95
& 15.19 & 0.4156 & 23.64 \\

DanceOPD (Single+Joint)
& 0.9236 & \underline{11.19} & 0.4123 & 23.73
& 0.8875 & \textbf{14.09} & 0.4604 & \textbf{23.44}
& \underline{15.27} & 0.4162 & 23.70 \\

\midrule
\textbf{CapField-OPD} (Single mode)
& \textbf{0.9531} & 8.50 & 0.4170 & 23.02
& \textbf{0.9592} & 11.96 & 0.4439 & 22.51
& \textbf{15.32} & \textbf{0.4182} & \underline{23.71} \\

\textbf{CapField-OPD} (Joint mode)
& 0.9322 & \textbf{11.76} & \textbf{0.4201} & \textbf{23.94}
& 0.9256 & \underline{14.08} & 0.4538 & 23.32
& -- & -- & -- \\

\bottomrule
\end{tabular}%
}
\vspace{-0.8em}
\end{table}

\begin{figure*}[t]
    \centering
    \includegraphics[width=\textwidth]{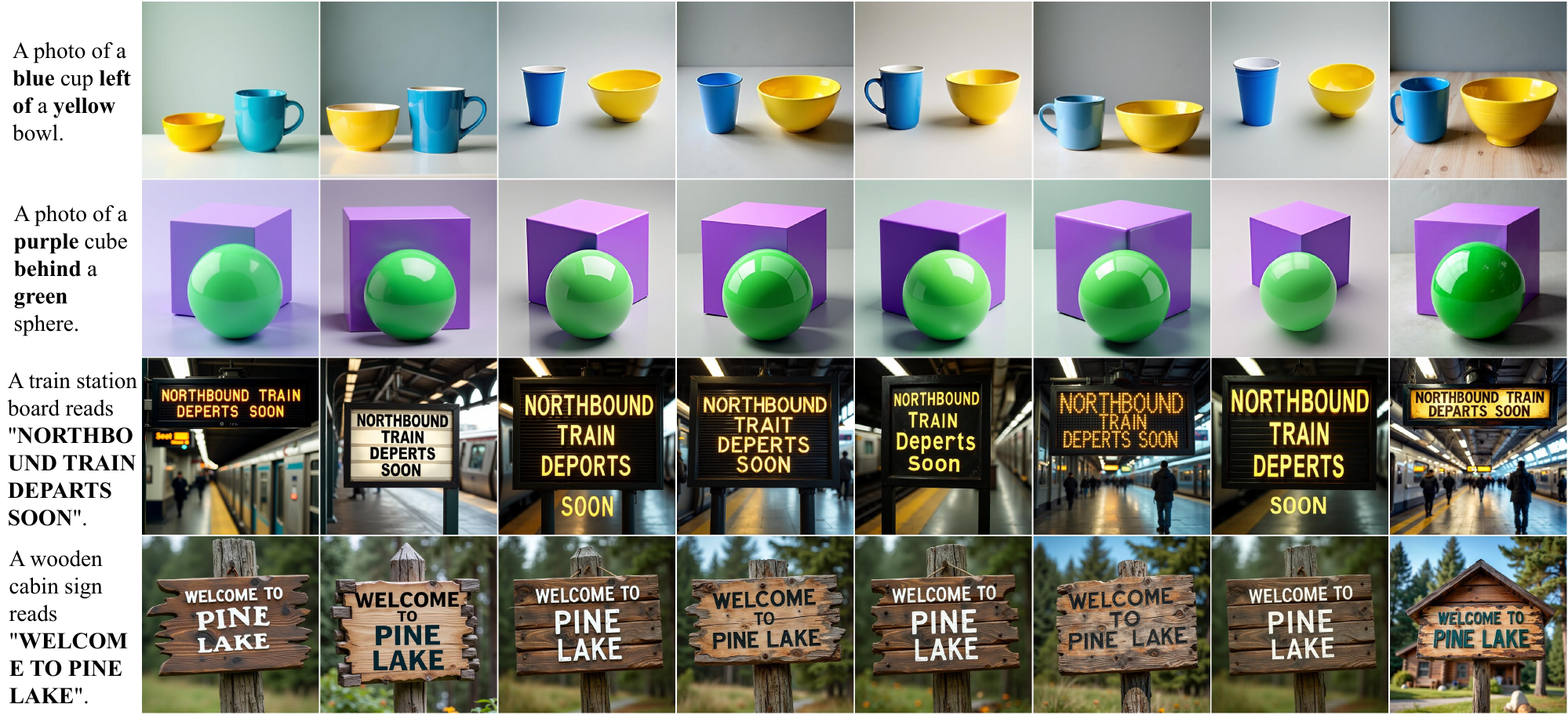}
    \vspace{-2em}
    \caption{
        Visual demonstration of different methods, from left to right: FLUX.1-dev, Multi-task GRPO, DiffusionOPD(Single-only), DiffusionOPD(Single+Joint), DanceOPD(Single-only), DanceOPD(Single+Joint), CapField-OPD(Single mode), and CapField-OPD(Joint mode).
    }
    \label{fig:visual_demonstration}
    \vspace{-1.2em}
\end{figure*}

\subsection{Main results}
Tab.~\ref{tab:main_results} presents the main quantitative comparison. The single-capability teachers perform strongly on their target tasks, while the joint teachers provide better balance between correctness and visual quality. Multi-task GRPO and existing OPD methods consolidate these capabilities into one model with a single operating point, and adding joint teachers (Single+Joint) lifts several aesthetic metrics but consistently lowers the GenEval and OCR scores. Without teacher identity as input, this supervision shifts the learned compromise rather than creating separately accessible capability modes. In comparison, the two CapField-OPD rows come from the same student model and differ only in capability coordinates. Single-capability coordinates achieve the best GenEval and OCR scores overall and slightly exceed the aesthetic teacher on all three aesthetic metrics, and some of these coordinates lie beyond the training anchors, so the anchors do not limit the learned field. Joint-capability coordinates give better-balanced operating points while retaining strong task accuracy. CapField-OPD thus supports capability extrapolation and direct switching between specialized and joint behaviors in one model, without retraining. Fig.~\ref{fig:visual_demonstration} shows the same trend qualitatively: in single mode, CapField-OPD matches the strongest baseline on text rendering, and joint mode improves visual quality while keeping the text correct.

\begin{table}[t]
    \centering
    \vspace{-1em}
    \caption{
    Robustness evaluation under semantics-preserving prompt rewriting.
    }
    \label{tab:prompt_robustness}

    \small
    \setlength{\tabcolsep}{3.0pt}
    \renewcommand{\arraystretch}{1.12}

    \begin{tabular*}{1.0\textwidth}{
        @{\extracolsep{\fill}}lccccccc@{}
    }
        \toprule
        \multirow{2}{*}{Metric}
        & \multirow{2}{*}{FLUX.1-dev}
        & \multirow{2}{*}{\shortstack{Multi-task\\GRPO}}
        & \multicolumn{2}{c}{DiffusionOPD}
        & \multicolumn{2}{c}{DanceOPD}
        & \multirow{2}{*}{\shortstack{CapField-OPD\\Single mode}} \\
        \cmidrule(lr){4-5}
        \cmidrule(lr){6-7}
        &
        &
        & Single
        & Single+Joint
        & Single
        & Single+Joint
        & \\
        \midrule

        GenEval $\uparrow$
        & 0.6550
        & 0.8461
        & 0.6241
        & 0.6132
        & 0.6838
        & 0.5560
        & \textbf{0.9267} \\

        OCR $\uparrow$
        & 0.6115
        & 0.9242
        & 0.8662
        & 0.8718
        & 0.8879
        & 0.8097
        & \textbf{0.9677} \\

        HPSv3 $\uparrow$
        & 13.46
        & 14.45
        & 15.13
        & 15.15
        & 15.08
        & 15.17
        & \textbf{15.28} \\

        \bottomrule
    \end{tabular*}
    \vspace{-1em}
\end{table}

\subsection{Robustness to Prompt Rewriting}
\label{sec:prompt_robustness}

We rewrite the benchmark prompts with GPT-5.6~\citep{openai2026gpt56systemcard} while preserving their task content and evaluate all methods without retraining. As shown in Tab.~\ref{tab:prompt_robustness}, CapField-OPD achieves the best result on all three metrics, while the compared OPD methods degrade significantly, especially in the GenEval and OCR tasks. This is because these methods infer the capability from prompt semantics, so the rewriting operator perturbs the inferred capability. CapField-OPD instead invokes it through an explicit coordinate that does not change with the wording, decoupling control from prompt phrasing: the learned capabilities transfer to new prompt types and styles.

\begin{figure*}[t]
    \centering
    \includegraphics[width=\textwidth]{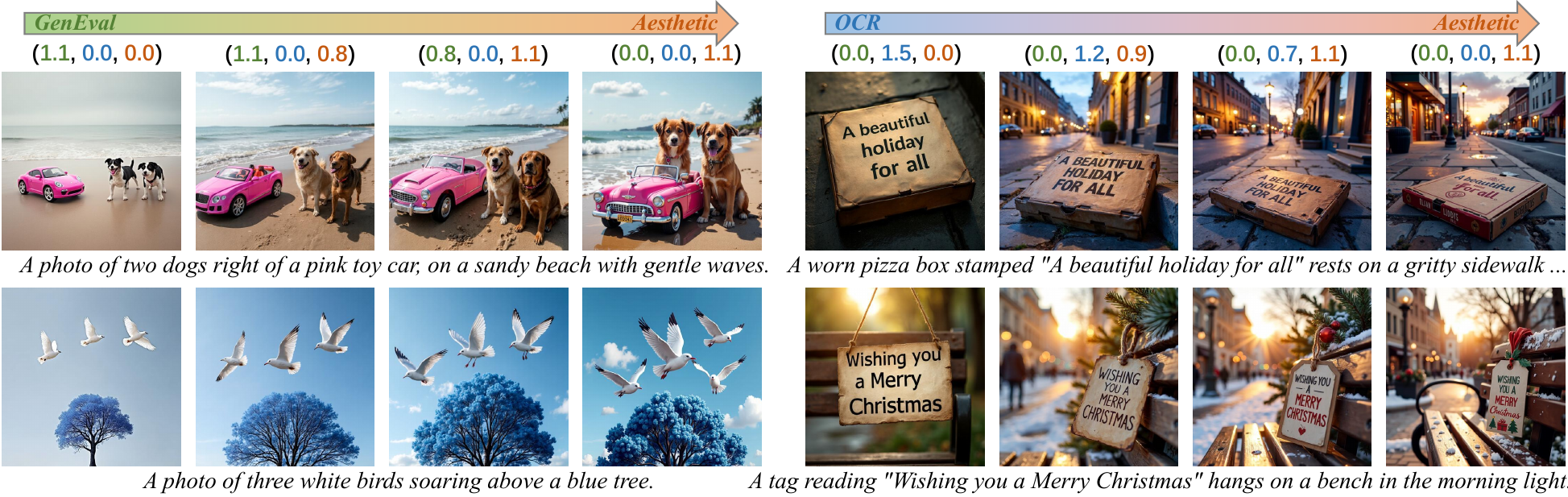}
    \vspace{-2em}
    \caption{
        Illustration of continuous capability control at inference time.
    }
    \label{fig:continuous_control}
    \vspace{-1em}
\end{figure*}

\subsection{Continuous Capability Control}

As shown in Fig.~\ref{fig:continuous_control}, sweeping the capability
coordinates with the prompt and initial noise fixed produces smooth
transitions between task-oriented and aesthetic behaviors.
Moving from GenEval toward aesthetics gradually enriches scene details,
whereas moving from OCR toward aesthetics changes lighting and visual
style while text fidelity is strongest near the OCR endpoint.
The main subjects and overall content remain recognizable throughout
each sweep, showing that the coordinates adjust the relative capability
emphasis without abrupt behavior changes.

\begin{figure*}[t]
    \centering
    \includegraphics[width=\textwidth]{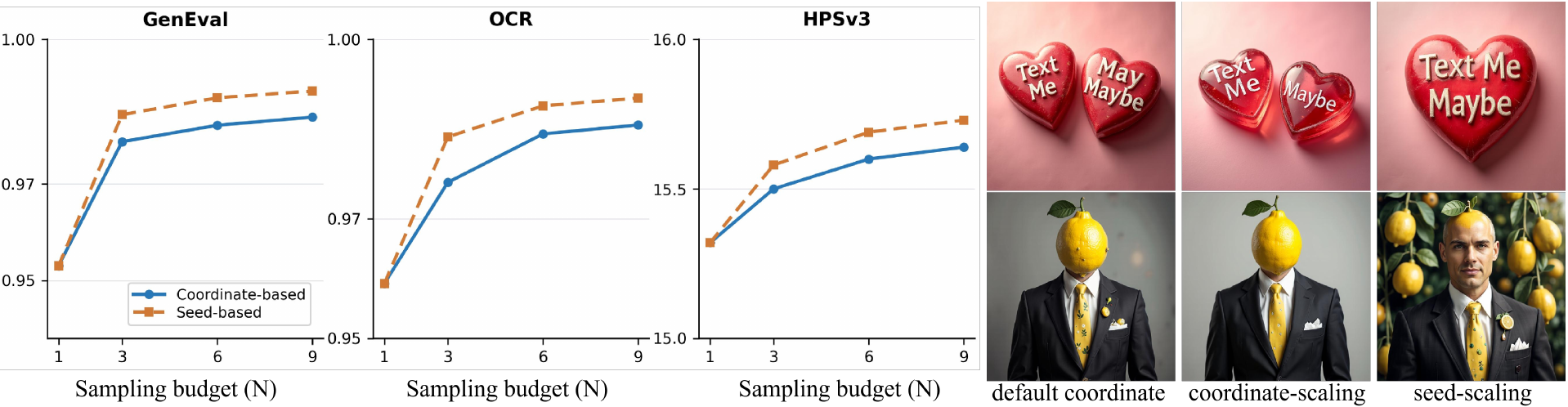}
    \vspace{-2em}
    \caption{
        Demonstration of coordinate-based and seed-based test-time scaling.
    }
    \label{fig:tts_visualization}
    \vspace{-1em}
\end{figure*}

\subsection{Test-Time Scaling over Capability Coordinates}
\label{sec:tts_experiments}

We compare coordinate-based scaling with conventional seed-based
scaling under the same generation budget $N\in\{3,6,9\}$.
Seed-based scaling fixes the capability configuration and resamples
the initial noise, while coordinate-based scaling fixes the noise and
searches over capability coordinates.
As shown in Fig.~\ref{fig:tts_visualization}, coordinate-based scaling improves the reward steadily as the budget grows and matches seed-based scaling across the three reward objectives, with a difference of about $0.5\%$.
Moreover, seed-based scaling gains reward by resampling the noise, so a
higher-reward candidate often comes with a different scene layout or
visual appearance. In comparison, coordinate-based scaling keeps the noise fixed and adjusts only
capability strengths, so task errors such as incorrect text or wrong
object counts are corrected while the original scene is largely
preserved.
Coordinate search is therefore preferable when the initial generation is largely satisfactory and only minor capability defects remain to be fixed.

\begin{table*}[t]
\centering
\caption{
Effect of joint-teacher anchors at jointly activated tasks.
}
\label{tab:joint_teacher_ablation}
\small
\setlength{\tabcolsep}{3.5pt}
\renewcommand{\arraystretch}{1.12}

\resizebox{\textwidth}{!}{%
\begin{tabular}{lcccccccc}
\toprule
\multirow{2}{*}{Training anchors}
& \multicolumn{4}{c}{Composition + Aesthetics}
& \multicolumn{4}{c}{Text Rendering + Aesthetics} \\
\cmidrule(lr){2-5}
\cmidrule(lr){6-9}
& GenEval $\uparrow$
& HPSv3 $\uparrow$
& CLIP $\uparrow$
& PickScore $\uparrow$
& OCR $\uparrow$
& HPSv3 $\uparrow$
& CLIP $\uparrow$
& PickScore $\uparrow$ \\
\midrule
Base + single
& 0.7723 & 9.38 & 0.4034 & 22.23
& 0.8716 & 13.50 & 0.4481 & 22.84 \\

Base + single + joint
& \textbf{0.9322} & \textbf{11.76} & \textbf{0.4201} & \textbf{23.94}
& \textbf{0.9256} & \textbf{14.08} & \textbf{0.4538} & \textbf{23.32} \\
\bottomrule
\end{tabular}%
}
\vspace{-1em}
\end{table*}

\begin{table*}[!t]
\centering
\caption{
Effect of coordinate extrapolation. Each $\hat{\lambda}^S_{\mathrm{peak}}$ (Eq.~\ref{eq:peak_and_search})
reports the full selected coordinate over (GenEval, OCR, aesthetics), with $S$ being
the single capability of the corresponding metric. }
\label{tab:coordinate_extrapolation_ablation}
\small
\setlength{\tabcolsep}{3.5pt}
\renewcommand{\arraystretch}{1.12}

\resizebox{\textwidth}{!}{%
\begin{tabular}{lcccccccc}
\toprule
\multirow{2}{*}{Coordinate selection}
& \multicolumn{2}{c}{Composition}
& \multicolumn{2}{c}{Text Rendering}
& \multicolumn{4}{c}{Aesthetics} \\
\cmidrule(lr){2-3}
\cmidrule(lr){4-5}
\cmidrule(lr){6-9}
& $\hat{\lambda}^S_{\mathrm{peak}}$
& GenEval $\uparrow$
& $\hat{\lambda}^S_{\mathrm{peak}}$
& OCR $\uparrow$
& $\hat{\lambda}^S_{\mathrm{peak}}$
& HPSv3 $\uparrow$
& CLIP $\uparrow$
& PickScore $\uparrow$ \\
\midrule
Teacher model
& / & 0.9417
& / & 0.9417
& / & 15.23 & 0.4181 & 23.68 \\

In-range profiling
& (1.0, 0.0, 0.0) & 0.9487
& (0.0, 1.0, 0.0) & 0.9390
& (0.0, 0.0, 1.0) & 15.25 & 0.4171 & 23.70 \\

Expanded profiling
& (1.05, 0.0, 0.0) & \textbf{0.9531}
& (0.0, 1.5, 0.0) & \textbf{0.9592}
& (0.0, 0.0, 1.3) & \textbf{15.32} & \textbf{0.4182} & \textbf{23.71} \\
\bottomrule
\end{tabular}%
}
\vspace{-1em}
\end{table*}

\subsection{Ablation Studies}

\paragraph{Effect of explicit capability coordinates.}
Tab.~\ref{tab:main_results} illustrates the limitation of consolidating
teacher behaviors into a single operating point.
For DiffusionOPD and DanceOPD, adding joint teachers improves
aesthetic metrics but reduces task accuracy, because the student receives no capability signal
and one prompt can correspond to several plausible targets.
CapField-OPD resolves this ambiguity by making capability intent explicit through coordinates, enabling a single student to support both specialized and joint behaviors.

\paragraph{Effect of joint teachers.}
Tab.~\ref{tab:joint_teacher_ablation} isolates the contribution of joint
anchors under the same conditioning architecture.
Adding joint teachers increases GenEval from $0.7723$ to $0.9322$ and
OCR from $0.8716$ to $0.9256$ under joint activation, while improving
all three aesthetic metrics for both capability pairs.
Thus, single-capability anchors alone do not fully determine the behavior
required at joint coordinates; joint teachers directly anchor these
combined operating points.

\paragraph{Effect of coordinate extrapolation.}
As shown in Tab.~\ref{tab:coordinate_extrapolation_ablation}, expanded profiling improves all metrics over in-range profiling, and the selected coordinates lie beyond the training range. This improvement arises from the continuity of the learned field: it maps coordinate changes to response changes, so the trend observed between anchors extends past them. The training anchors therefore define reference states rather than the best inference settings of the learned field.

\subsection{Limitation}
Although the training target is constructed by fusing multiple teacher outputs, these outputs do not participate in rollout and require no gradients, so the extra cost is limited. In practice, compared to DiffusionOPD under the same backbone, batch size, and sampling steps, the per-step training time increases only from 11 s to 12 s (about $9\%$), which comes from the additional frozen-teacher forward passes used to build the joint-anchored target.
Additionally, the proposed method builds the student by consolidating the
capabilities of the teachers. If a teacher trained in the RL stage
suffers from problems such as reward hacking, the student may inherit
similar behaviors from its supervision.

\section{Conclusion}
In this work, we present CapField-OPD, an on-policy distillation framework that learns a continuous capability field over explicit capability coordinates, where reward-specialized teachers serve as anchors. Existing multi-teacher OPD methods infer the desired capability from prompt content, which binds capability control to prompt wording and prevents users from adjusting it at inference time. By conditioning the student on these coordinates, CapField-OPD decouples capability control from prompt semantics. The student thus matches or exceeds every anchor on its target task, and the requested capability stays active under prompt rewriting. The coordinates can also extend beyond the training anchors, which serve as reference states rather than hard performance limits. More broadly, explicit capability coordinates offer a way to build generative models whose behaviors are specified by users rather than inferred from prompts.

\newpage

\bibliography{iclr2027_conference}
\bibliographystyle{iclr2027_conference}


\newpage
\appendix

\section{Appendix}
This appendix provides the training pseudocode of CapField-OPD
(Sec.~\ref{app:pseudo_code}), the full hyperparameter configuration
together with the capability curves during distillation
(Sec.~\ref{app:hyperparameters}), and additional visualizations of
coordinate-based strength control (Sec.~\ref{app:control_vis}).

\subsection{Pseudocode}
\label{app:pseudo_code}

Algorithm~\ref{alg:capfield_opd} summarizes CapField-OPD training. The
single-capability and joint teachers are obtained by reward-based
post-training and then frozen, and they supervise the student only at
student-visited states. Each prompt $c$ is paired with a set
$\Lambda(c)$ of matching capability coordinates, and all coordinate
sampling is tied to this set. Each rollout is conditioned on a
coordinate $\boldsymbol{\lambda}^{\rm roll}$ drawn from $\Lambda(c)$,
and the capability coordinates can be different when computing OPD loss. A
single rollout thus supervises many coordinates, so the rollout cost
does not scale with the number of supervised coordinates. For pairs
without a joint teacher, both the rollout and the supervised
coordinates are restricted to axis-aligned ones
(Sec.~\ref{sec:capability_field}).

\begin{algorithm*}[ht]
    \caption{CapField-OPD Training}
    \label{alg:capfield_opd}
    \footnotesize
    \begin{algorithmic}
        \vspace{-0.2em}

        \Statex \textbf{Input:} Frozen base teacher $v_0$, single and
        joint teachers $\{v_i\},\{v_{ij}\}$; prompt datasets
        $\{\mathcal{D}_k\}$, where each prompt $c$ is paired with its
        matching capability coordinates $\Lambda(c)$; noise schedule
        $\{t_r\}_{r=0}^{S}$.

        \Statex \textbf{Output:} Student
        $v_\theta(\mathbf{x},t,c,\boldsymbol{\lambda})$.

        \State Initialize $v_\theta$ from $v_0$ with a trainable LoRA;
        add the coordinate conditioner
        (Eq.~\ref{eq:capability_conditioning}) with
        $W_2\gets\mathbf{0}$.

        \For{each training iteration}

            \State Sample a balanced batch of prompts
            $c\sim\mathcal{D}_k$; for each prompt, draw a rollout
            coordinate
            $\boldsymbol{\lambda}^{\rm roll}\sim\Lambda(c)$.

            \State Roll out the student on each $(c,
            \boldsymbol{\lambda}^{\rm roll})$ to obtain on-policy
            trajectories $\{\mathbf{x}_{t_r}\}_{r=0}^{S}$.
            \Comment{no gradient}

            \For{each intermediate state
            $(\mathbf{x}_{t_r},t_r,c)$}

                \State Draw a capability coordinate
                $\boldsymbol{\lambda}\sim\Lambda(c)$ for loss computation.

                \State Compute the teacher target $\mathbf{v}^\star$
                via Eq.~\ref{eq:capability_field} at
                $(\mathbf{x}_{t_r},t_r,c)$.
                \Comment{no gradient}

                \State Accumulate the squared error
                $\|v_\theta(\mathbf{x}_{t_r},t_r,c,
                \boldsymbol{\lambda})-\mathbf{v}^\star\|_2^2$
                into $\mathcal{L}_{\rm OPD}$.

            \EndFor

            \State Update the LoRA and the coordinate conditioner based on
            the mean of $\mathcal{L}_{\rm OPD}$.

        \EndFor

        \vspace{-0.2em}
    \end{algorithmic}
\end{algorithm*}

\subsection{Hyperparameter Configuration and Capability Curves}
\label{app:hyperparameters}

\begin{figure*}[ht]
    \centering
    \includegraphics[width=\textwidth]{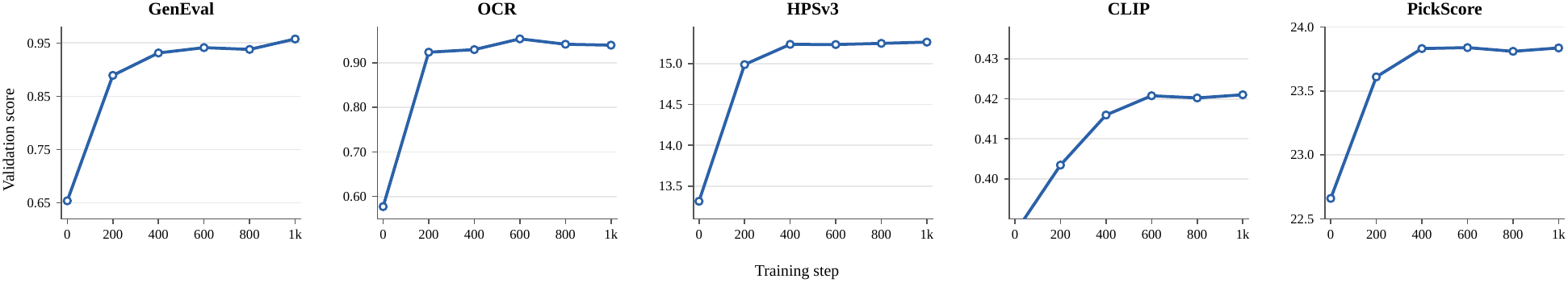}
    \vspace{-2em}
    \caption{Capability curves during CapField-OPD training. }
    \label{fig:trainingcurve}
    \vspace{-1em}
\end{figure*}

Table~\ref{tab:capfield_hyperparameters} lists the hyperparameter
settings used for CapField-OPD training.
Fig.~\ref{fig:trainingcurve} reports the validation reward of each
capability over distillation steps, evaluated every 200 steps on
validation prompts per capability set, in which 
all the capabilities improve steadily during training.

\begin{table}[ht]
    \centering
    \caption{Hyperparameter settings used for CapField-OPD.}
    \label{tab:capfield_hyperparameters}

    \footnotesize
    \setlength{\tabcolsep}{2.4pt}
    \renewcommand{\arraystretch}{1.08}

    \begin{tabular*}{\columnwidth}{
        @{\extracolsep{\fill}}lclc@{}
    }
        \toprule
        \textbf{Parameter}
        & \textbf{Value}
        & \textbf{Parameter}
        & \textbf{Value} \\
        \midrule

        Random seed
        & $42$
        & Learning rate
        & $2\!\times\!10^{-4}$ \\

        Optimizer
        & AdamW
        & Weight decay
        & $1\!\times\!10^{-4}$ \\

        AdamW $\beta_1,\beta_2$
        & $0.9,\,0.999$
        & AdamW $\epsilon$
        & $1\!\times\!10^{-8}$ \\

        LR scheduler
        & Constant
        & Text length (CLIP / T5)
        & $77$ / $256$ \\

        LoRA rank / $\alpha$
        & $64$ / $128$
        & Cond. dimensions
        & $3\!\rightarrow\!256\!\rightarrow\!3072$ \\

        Batch size
        & $8$
        & Number of GPUs
        & $8$ \\

        Mixed precision
        & bfloat16
        & Max. grad norm
        & $1.0$ \\

        Denoising steps
        & $10$
        & Train time indices
        & $0,\ldots,9$ \\

        Resolution
        & $512\!\times\!512$
        & Guidance scale
        & $4.5$ \\

        Loss target
        & Velocity
        & Timestep aggregation
        & Mean \\

        EMA decay
        & $0.9$
        & EMA update interval
        & $8$ steps \\

        Calibration size $M$
        & $200$
        & Probe grid spacing
        & $0.05$ \\

        Extrapolation bound
        & $1.5$
        & Search budget $N$
        & $\{3,6,9\}$ \\

        \bottomrule
    \end{tabular*}
\end{table}

\subsection{Additional Visualizations of Capability Control}
\label{app:control_vis}

\begin{figure*}[t]
    \centering
    \includegraphics[width=0.85\textwidth]{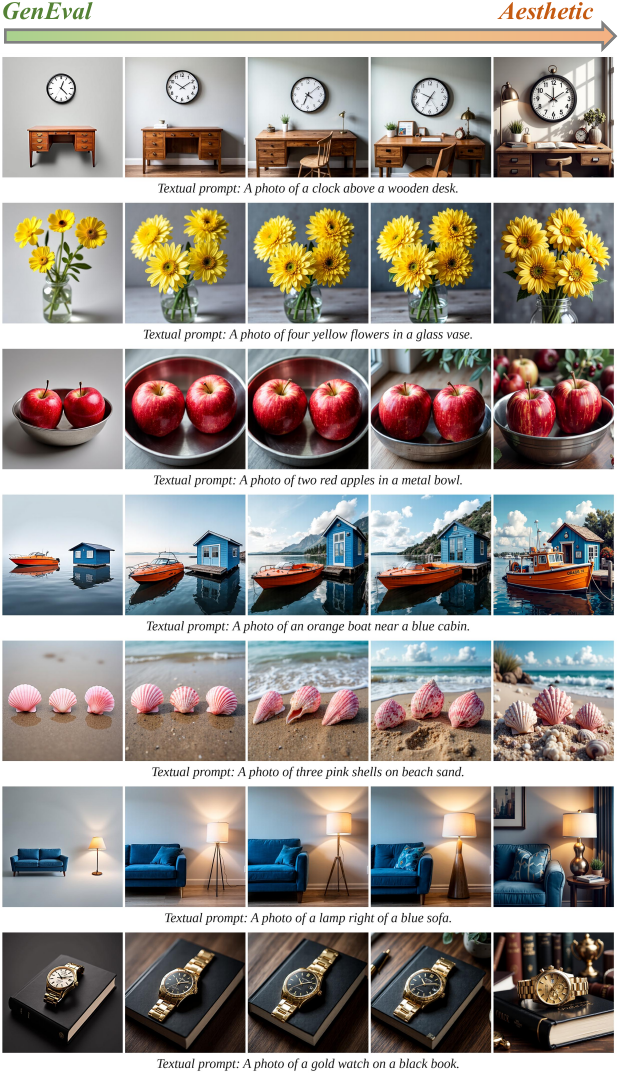}
    \vspace{-1em}
    \caption{
        Additional coordinate sweeps at inference time from GenEval to aesthetics.
    }
    \label{fig:continuous_control_geneval}
\end{figure*}

\begin{figure*}[t]
    \centering
    \includegraphics[width=0.85\textwidth]{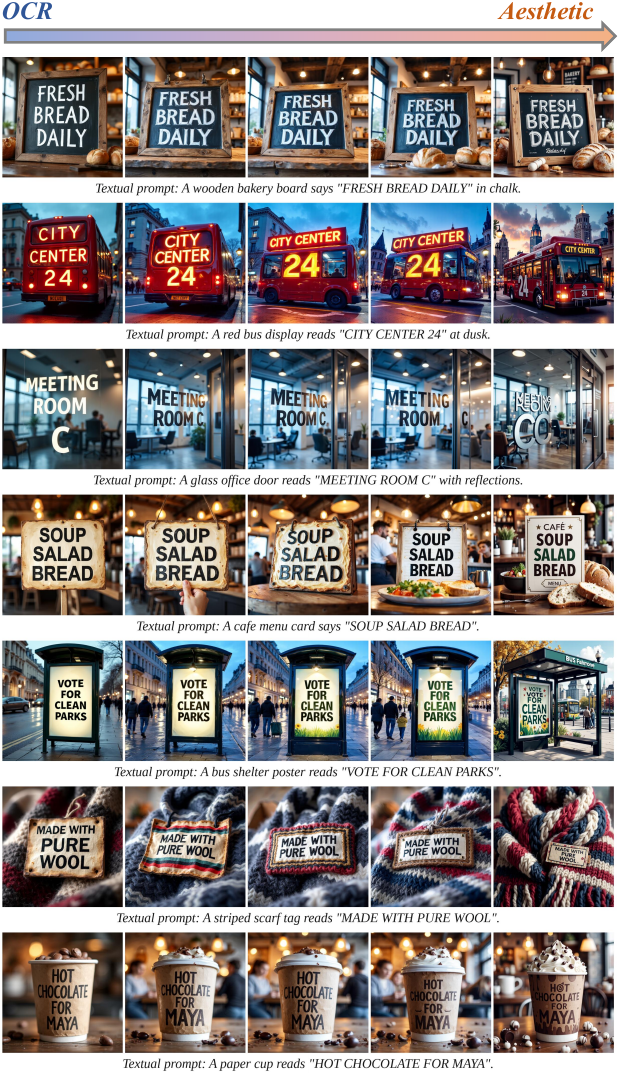}
    \vspace{-1em}
    \caption{
        Additional coordinate sweeps at inference time from OCR to aesthetics.
    }
    \label{fig:continuous_control_ocr}
\end{figure*}

Fig.~\ref{fig:continuous_control_geneval} and Fig.~\ref{fig:continuous_control_ocr} represent more visual results of coordinate sweeping at inference time for the two supported pairs, (GenEval, aesthetics) and (OCR, aesthetics). Each row fixes the prompt and the initial noise and varies only the coordinates, so the differences within a row come from the capability field rather than the seed. For a given prompt-noise pair, sweeping the coordinate moves the output smoothly along the corresponding capability axis while the composition stays close to the shared starting point. The control behavior is therefore a property of the field itself, not of a particular capability pair or seed.

\end{document}